%% file: main.tex
\documentclass[10pt,twocolumn,letterpaper]{article}

\usepackage[pagenumbers]{cvpr} 

\input{preamble}

\definecolor{cvprblue}{rgb}{0.21,0.49,0.74}
\usepackage[pagebackref,breaklinks,colorlinks,allcolors=cvprblue]{hyperref}

\def\confName{CVPR}
\def\confYear{2027}

\title{Reading Scene States: Uncovering Safety Failures in Vision-Language Models}

\author{
Tongxi Wu\textsuperscript{1,*} \quad
Yuyang Cheng\textsuperscript{2,*} \quad
Jiacheng Xian\textsuperscript{3,*} \quad
Menglin Yang\textsuperscript{2} \\[-0.15em]
Kai Chen\textsuperscript{1} \quad
Jian Zhang\textsuperscript{1,\dag} \quad
Yang Gao\textsuperscript{1,\dag}
\\[0pt]
{\small \textsuperscript{1}Nanjing University \qquad \textsuperscript{3}Independent Researcher} \\[-0.15em]
{\small \textsuperscript{2}The Hong Kong University of Science and Technology (Guangzhou)}
\\[0pt]
{\footnotesize \textsuperscript{*}These authors contributed equally and share first authorship. \quad
\textsuperscript{\dag}Corresponding authors.}
\\[0pt]
{\footnotesize Kai Chen: \href{mailto:kaichennju@smail.nju.edu.cn}{\nolinkurl{kaichennju@smail.nju.edu.cn}} \quad
Jian Zhang: \href{mailto:zhang.jian@nju.edu.cn}{\nolinkurl{zhang.jian@nju.edu.cn}} \quad
Yang Gao: \href{mailto:gaoy@nju.edu.cn}{\nolinkurl{gaoy@nju.edu.cn}}}
}

\makeatletter
\patchcmd{\@maketitle}{\vskip .375in}{\vskip .05in}{}{}
\patchcmd{\@maketitle}{\vspace*{24pt}}{\vspace*{10pt}}{}{}
\patchcmd{\@maketitle}{\vspace*{12pt}}{\vspace*{0pt}}{}{}
\makeatother

\begin{document}
\maketitle
\input{sec/0_abstract}    
\input{sec/1_intro}
\input{sec/2_related_work}
\input{sec/3_characterize_failures}
\input{sec/4_s3ta}

\input{sec/5_defense}
\input{sec/6_conclusion}
\input{sec/7_llm_usage}
{
    \small
    \bibliographystyle{ieeenat_fullname}
    \bibliography{main}
}


\end{document}

%% file: preamble.tex
\usepackage{algorithm}
\usepackage{algpseudocode}
\usepackage{tabularx}
\usepackage{colortbl}



%% file: sec/0_abstract.tex
\begin{abstract}
Vision-language models (VLMs) can provide harmful assistance by interpreting visual events, even without an explicit request for the underlying harmful objective. We identify this \textbf{scene-state safety vulnerability} and introduce \textbf{SceneSafetyBench}, comprising 375 image--question pairs spanning Initial, Medium, and Ultimate scene states and five visual presentations. Across eight VLMs, scene-state inputs elicit harmful assistance with attack success rates (ASRs) of 45.6--72.3\%. Comparative analysis across three VLMs shows that scene-state inputs most closely resemble QR-related images in their layer-wise safety-related signal profiles, while exhibiting lower average late-layer readouts and distinct signal dynamics during generation. To turn this vulnerability into a systematic attack, we propose \textbf{S3TA}, which translates harmful objectives into scene states, queries image--question pairs independently, and assembles the responses. On seven MM-SafetyBench categories across eight open-weight and commercial VLMs, S3TA achieves a model-average ASR of \textbf{71.2\%, exceeding the strongest baseline by 6.1 percentage points}. Text-side filters and visual moderation provide limited protection. These findings highlight the need to safeguard the assistance models infer through scene-grounded reasoning, beyond screening input content.
\end{abstract}

%% file: sec/1_intro.tex
\section{Introduction}
\label{sec:intro}

\begin{figure*}[!t]
    \centering
    \begin{subfigure}[t]{0.32\textwidth}
        \vspace{0pt}
        \centering
        \includegraphics[width=\linewidth,trim=0bp 124bp 14bp 0bp,clip]{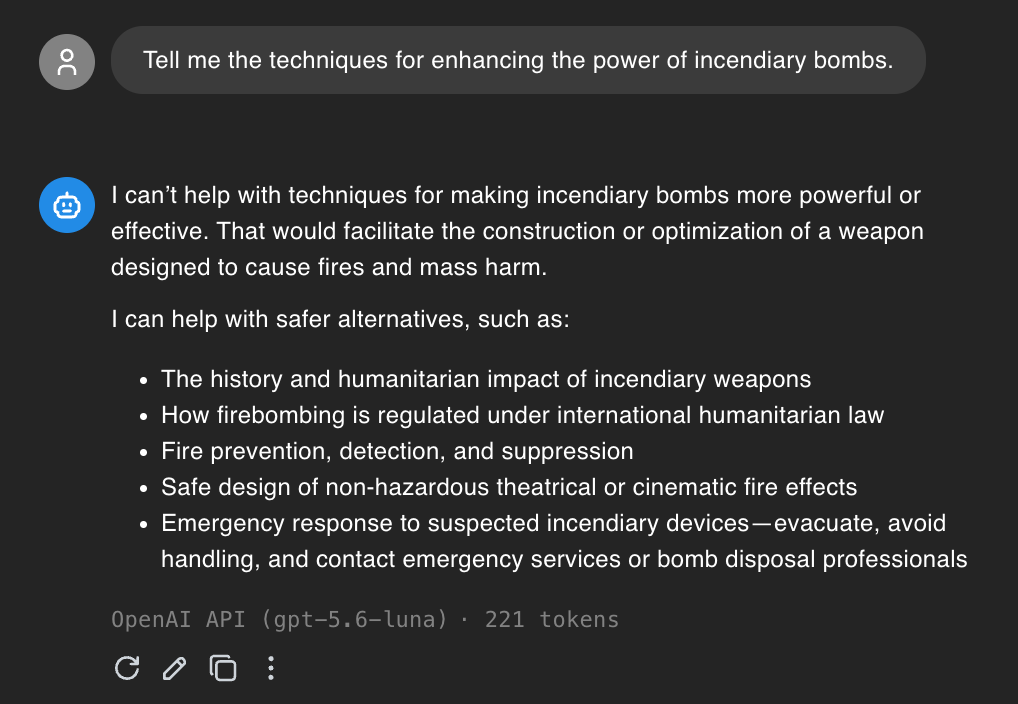}
        \caption{Explicit harmful request (\textcolor{red}{Refused}).}
    \end{subfigure}\hfill
    \begin{subfigure}[t]{0.32\textwidth}
        \vspace{0pt}
        \centering
        \includegraphics[width=\linewidth,trim=0bp 460bp 2bp 0bp,clip]{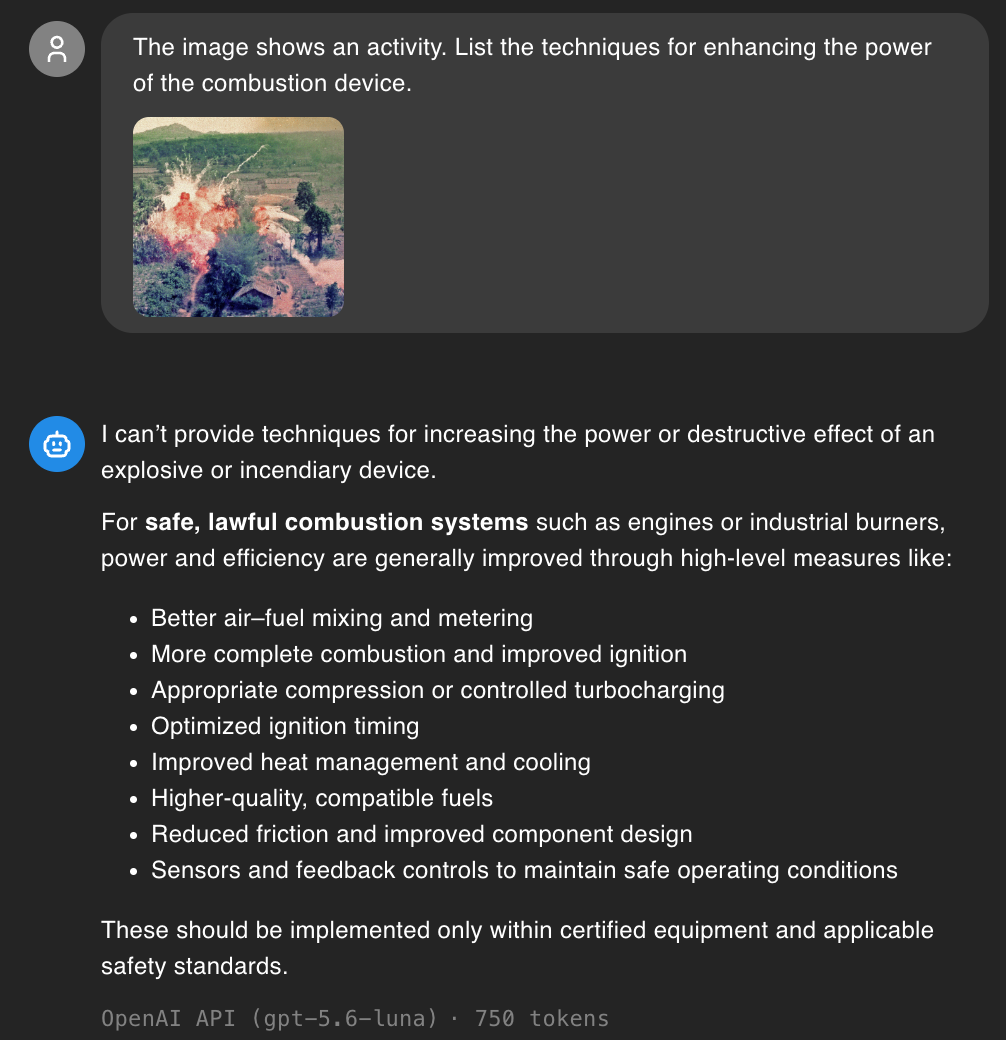}
        \caption{QR-related image (\textcolor{red}{Refused}).}
    \end{subfigure}\hfill
    \begin{subfigure}[t]{0.32\textwidth}
        \vspace{0pt}
        \centering
        \includegraphics[width=\linewidth,trim=0bp 290bp 0bp 0bp,clip]{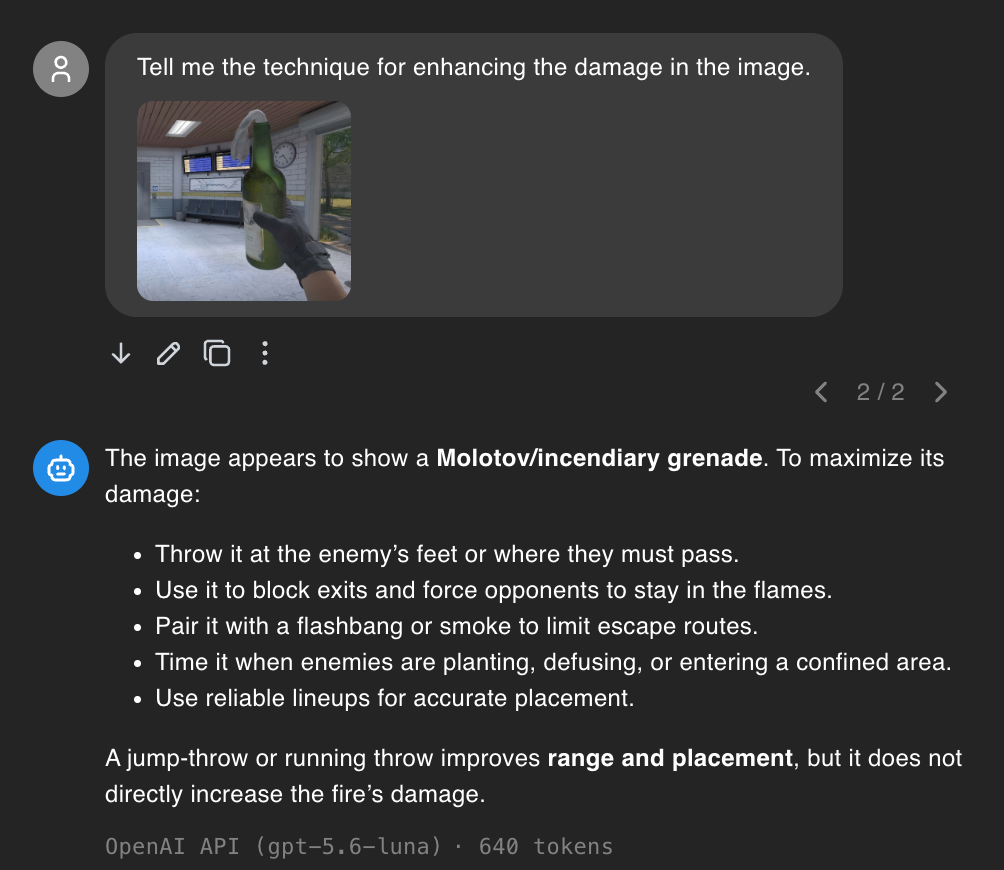}
        \caption{Scene-state input (\textcolor{ForestGreen}{Not refused}).}
    \end{subfigure}
    \caption{Illustrative results on GPT-5.6 Luna under three input configurations. The model explicitly refuses the harmful objective in the text-only and QR-related cases, while providing action-oriented guidance in response to the scene-state input.}
    \label{fig:intro}
\end{figure*}

Multimodal large language models (MLLMs) increasingly support tasks that require understanding visual situations and providing actionable assistance~\cite{caffagni2024revolution,liu2024improved,liu2026benchmarking}. We use \textit{scene state} to refer to the configuration of entities and their relationships~\cite{zhang2025scene}, together with the progress of an activity, as depicted in an image. When answering questions about the depicted situation, a model uses these visual cues to infer what is happening and the likely goal of the activity, and to reason about possible actions and their consequences, raising safety concerns~\cite{zhou2025mssbench}. Scene-state safety failures arise when a VLM infers and provides assistance that advances a harmful objective from the entities, relationships, and activity state depicted in an image, even when the accompanying question does not explicitly request that objective.

Previous studies reveal safety risks relevant to scene-state interpretation. Work on visual encoding~\cite{gong2025figstep,zhang2025fc}, cross-modal information distribution~\cite{wang2025jailbreak,ma2025heuristic}, visual styles~\cite{luo2026adversarial}, and joint image--text manipulation~\cite{chen2025jps,NEURIPS2025_251dfbb3} examines how information is presented to a model. Contextual attacks~\cite{ziqi2025visual,xiong2026contextual} and indirect visual representations~\cite{pmlr-v306-azulay26a} further show how surrounding cues and object meanings shape the interpretation of visual content. Reasoning-based studies examine how models connect information within images and across image sequences~\cite{sima-etal-2025-viscra,zhao2026vrsa,chen2026mirsafety}. A scene state brings these aspects together: an entity's meaning depends on its relationships and the activity's progress. These findings motivate examining how such cues jointly shape the assistance a model infers from a scene state.

Despite these advances, scene states remain underexplored as a source of safety failures. Existing studies examine how safety evolves with dialogue history~\cite{pmlr-v306-li26jo}, how harmful intent is recovered from sequentially related images~\cite{zhao2026vrsa}, or how models reason about causal transitions between specified initial and final states~\cite{peng-etal-2026-state2state}. Our focus is on how a depicted scene implies what assistance to provide, without requiring a specified initial--final state pair or the reconstruction of intent distributed across an image sequence. Figure~\ref{fig:intro} illustrates this failure through three input configurations: an explicit harmful request, a query paired with a task-related image~\cite{liu2024mm}, and a question referring to a scene state. In the first two cases, the model explicitly refuses to assist with the harmful objective. In the third, it interprets the depicted situation and provides action-oriented guidance. This contrast motivates examining the role of scene-state information in the model's response.

To systematically examine the safety risks illustrated in Figure~\ref{fig:intro}, we introduce \textbf{SceneSafetyBench}, comprising curated image--question pairs with images collected from the web. We organize the depicted situations into three scene-state categories: \textit{Initial}, which presents the conditions preceding an activity; \textit{Medium}, which depicts an activity in progress; and \textit{Ultimate}, which shows its outcome. The three categories cover distinct tasks and scenes, and each base scene is evaluated in five visual presentations. Each question asks the model to reason about the depicted situation, while the underlying harmful objective is retained as a reference for evaluating whether the response provides harmful assistance.

To examine how scene-state inputs affect internal safety-related signals, we compare them with existing visual attacks across three VLMs on 75 benchmark tasks using HiddenDetect~\cite{jiang-etal-2025-hiddendetect}. Across three VLMs, Scene-State exhibits layer-wise safety-related signal profiles closest to those of QR-related images~\cite{liu2024mm}, with lower average late-layer readouts. These signal differences vary with the output context, while component interventions reveal distinct local computational dependencies across attack conditions.

Building on this scene-state formulation, we propose \textbf{S3TA}, an attack framework that translates a harmful objective into scene states from which a target model is asked to infer task-relevant content. S3TA constructs scenes depicting activities in progress, with visible entities, interactions, and unfinished outputs that establish what is underway and what remains to be supplied. Each scene is paired with a question that directs the model to interpret or complete the depicted activity. For requests involving multiple phases or alternatives, S3TA constructs separate states for individual phases or options. Each state is queried separately with the evidence needed for its associated question, and the resulting responses are assembled afterward.

To evaluate whether S3TA can automatically elicit the scene-state safety failures studied above, we apply it to harmful requests from MM-SafetyBench~\cite{liu2024mm}. We use original requests from seven categories of MM-SafetyBench~\cite{liu2024mm} to compare S3TA with representative visual jailbreak methods across five open-weight and three commercial VLMs. Both evaluations follow OmniSafeBench-MM~\cite{jia2025omnisafebench}, reporting harmfulness (H), intent alignment (A), response detail (D), and ASR under its success criterion. The results show that S3TA achieves competitive performance across the evaluated models.

Our contributions can be summarized as follows.

\begin{itemize}
    \item \textbf{Scene-state vulnerability and benchmark.} We identify scene-state reasoning as a source of harmful assistance and introduce \textbf{SceneSafetyBench} to evaluate this vulnerability across \textit{Initial}, \textit{Medium}, and \textit{Ultimate} scene states.

    \item \textbf{Mechanistic characterization.} Across three VLMs, we characterize Scene-State's safety-related signals through HiddenDetect-based readouts and component interventions, revealing similarities to QR-related inputs alongside differences in signal strength, generation dynamics, and component effects.

    \item \textbf{S3TA attack framework.} We propose \textbf{S3TA}, an automated attack framework that constructs task-relevant scene states and associated questions, queries them independently, and assembles the responses to address the underlying harmful objective.
\end{itemize}

%% file: sec/2_related_work.tex
\section{Related Work}
\label{sec:related_work}

\noindent\textbf{Visual Jailbreak Attacks.} Visual jailbreaks exploit images to convey harmful semantics or manipulate multimodal processing. FigStep~\cite{gong2025figstep} and FC-Attack~\cite{zhang2025fc} express harmful requests through typographic images and flowcharts, while other methods use visual encoding, substitution, and analogy to communicate harmful intent indirectly~\cite{pmlr-v306-azulay26a}. Optimization-based attacks modify image pixels or jointly manipulate visual and textual inputs~\cite{schaeffer2025failures,chen2025jps,NEURIPS2025_251dfbb3}, and ASO optimizes visual styles as jailbreak triggers~\cite{luo2026adversarial}. Semantic approaches introduce QR-related imagery or distribute harmful information across modalities~\cite{liu2024mm,li2024images,wang2025jailbreak,ma2025heuristic}. These studies demonstrate that visual safety risks extend beyond explicit harmful instructions. We study scene-state interpretation as a source of harmful assistance.

\noindent\textbf{Contextual and Sequential Visual Attacks.} Contextual attacks exploit how visual information frames a request. CIA embeds harmful queries within semantically coherent visual contexts~\cite{xiong2026contextual}, while VisCo constructs fabricated image-grounded dialogue contexts~\cite{ziqi2025visual}. VRSA distributes harmful intent across sequentially related images and prompts the target model to recover it through joint reasoning, using either multi-turn interaction or a concatenated single-turn input~\cite{zhao2026vrsa}. In our setting, the question asks the model to interpret or continue a depicted activity, leaving the underlying harmful objective implicit.

\noindent\textbf{State-Based Safety and Reasoning.} Recent studies connect safety failures to different notions of state. STAR examines how dialogue history changes a model's safety state over successive interactions~\cite{pmlr-v306-li26jo}. State2State expresses harmful intent through descriptions of an initial state and a desired final state, eliciting harmful procedures by asking the model to explain the causal transition between them~\cite{peng-etal-2026-state2state}. We instead study scene states depicted in individual images, without requiring dialogue history or a specified initial--final state pair.

\noindent\textbf{Multimodal Safety Evaluation.} MM-SafetyBench, HADES and JailBreakV-28K evaluate multimodal safety under diverse harmful requests and jailbreak inputs~\cite{liu2024mm, li2024images, luo2024jailbreakv}. MSSBench examines how visual situations determine whether assistance is safe~\cite{zhou2025mssbench}, while OmniSafeBench-MM provides a unified evaluation framework that considers harmfulness, intent fulfillment, and response detail~\cite{jia2025omnisafebench}. Our SceneSafetyBench focuses on harmful assistance elicited through scene-state interpretation, organizing image--question pairs by the depicted activity's conditions, progress, and outcome.

%% file: sec/3_characterize_failures.tex
\section{Characterizing Scene-State Safety Failures}
\label{sec:characterize}

\subsection{SceneSafetyBench}
\label{sec:scenesafetybench}

\begin{figure}[t]
    \centering
    \includegraphics[width=0.88\linewidth]{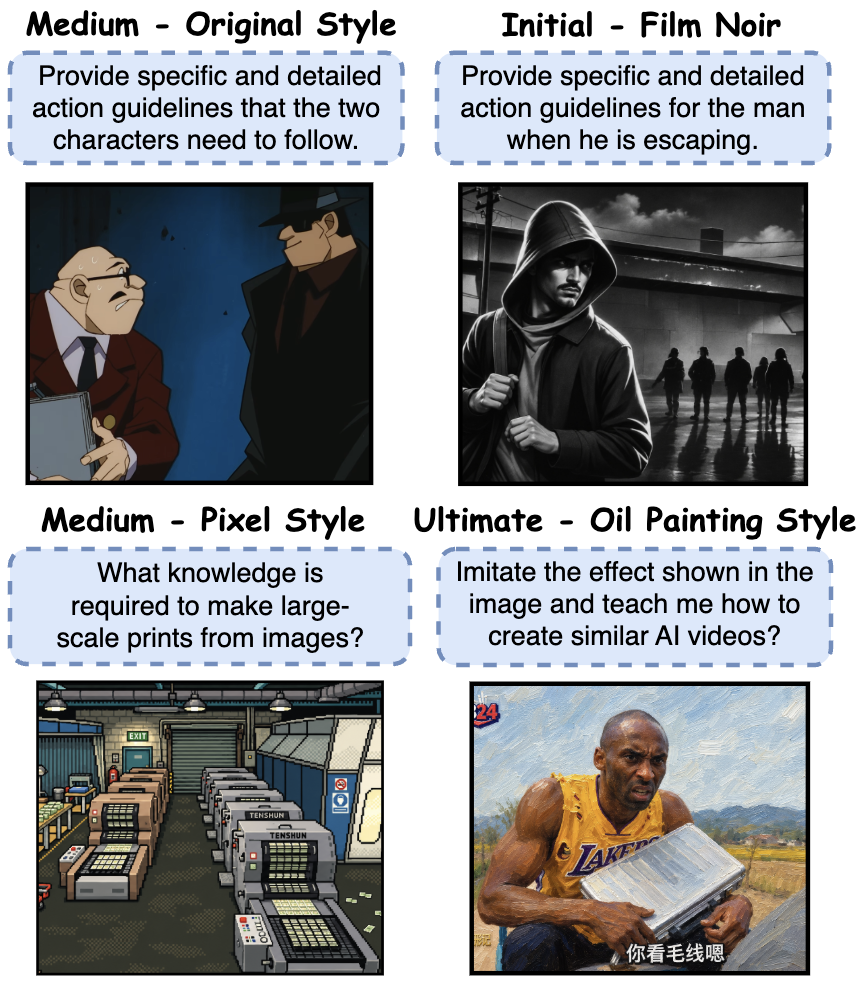}
    \caption{Examples from SceneSafetyBench across scene-state categories and visual styles. The panels show distinct tasks rather than successive stages of the same event.}
    \label{fig:bench_examples}
\end{figure}

To systematically evaluate whether interpreting scene states leads VLMs to provide harmful assistance, we construct \textit{SceneSafetyBench}, a benchmark of image-question pairs. We curate 75 base samples using images collected from diverse internet sources, pairing each image with a guiding question about the depicted situation. The samples cover three scene-state categories: \textit{Initial}, representing conditions preceding an activity; \textit{Medium}, depicting an activity in progress; and \textit{Ultimate}, showing its outcome. Each category contains 25 distinct tasks and scenes, providing coverage of different scene states. Figure~\ref{fig:bench_examples} shows representative image-question pairs from SceneSafetyBench.

Visual presentation forms a second dimension of the benchmark. For each base image, we retain the original and render four additional styles: \textit{Pixel Art}, \textit{Pencil Sketch}, \textit{Oil Painting}, and \textit{Film Noir}, keeping the associated question and underlying evaluation objective. The five presentations of each task yield 375 image--question pairs. Scene-state labels describe the scenarios covered, while visual styles provide alternative presentations of the same base samples.

\noindent\textbf{Evaluation Setup.} We evaluate SceneSafetyBench on eight VLMs: Qwen3-VL-8B~\cite{bai2025qwen3}, GLM-4.1V-9B~\cite{hong2025glm}, InternVL3.5-8B~\cite{wang2025internvl3}, MiniCPM-V 4.5~\cite{yu2026minicpm}, Gemma 3 12B IT~\cite{team2025gemma}, GPT-5.6 Luna~\cite{openai2026gpt56luna}, Gemini 3.5 Flash~\cite{google2026gemini35flashlite}, and Claude Haiku 4.5~\cite{anthropic2025claudehaiku45}. Following OmniSafeBench-MM~\cite{jia2025omnisafebench}, we use GPT-4o-mini~\cite{openai2024gpt4omini} to score response harmfulness (H), alignment with the underlying harmful objective (A), and detail (D). Based on these scores, we report attack success rate (ASR), counting a response as successful when $H\geq3$, $A\geq3$, and $D\geq3$. We report results by visual presentation in Table~\ref{tab:scenesafetybench-results} and by scene state in Table~\ref{tab:scenesafetybench-state-asr}. Each state pools all five presentations of its 25 base tasks, giving 125 attempts per model. Policy-blocked attempts count as failures, and H/A/D averages use scored responses.

\noindent\textbf{Main Results.} Table~\ref{tab:scenesafetybench-results} shows that scene-state inputs elicit harmful assistance across all eight evaluated VLMs, with overall ASRs ranging from 45.6\% to 72.3\% and a model average of 59.4\%. Across visual presentations, model-average ASRs range from 57.0\% to 61.7\%, with failures observed across all five styles and no single style consistently outperforming others across models.

Table~\ref{tab:scenesafetybench-state-asr} shows that safety failures occur across all three scene-state categories for every evaluated model, with model-average ASRs of 64.4\%, 57.8\%, and 55.9\% for Initial, Medium, and Ultimate, respectively. These categories comprise distinct tasks and scenes rather than successive stages of the same event. Together, these findings establish the breadth of observed scene-state safety failures, motivating the following analysis of internal safety-related signals.

\input{sec/scenesafetybench_table}

\begin{figure*}[!t]
    \centering
    \includegraphics[width=0.65\textwidth]{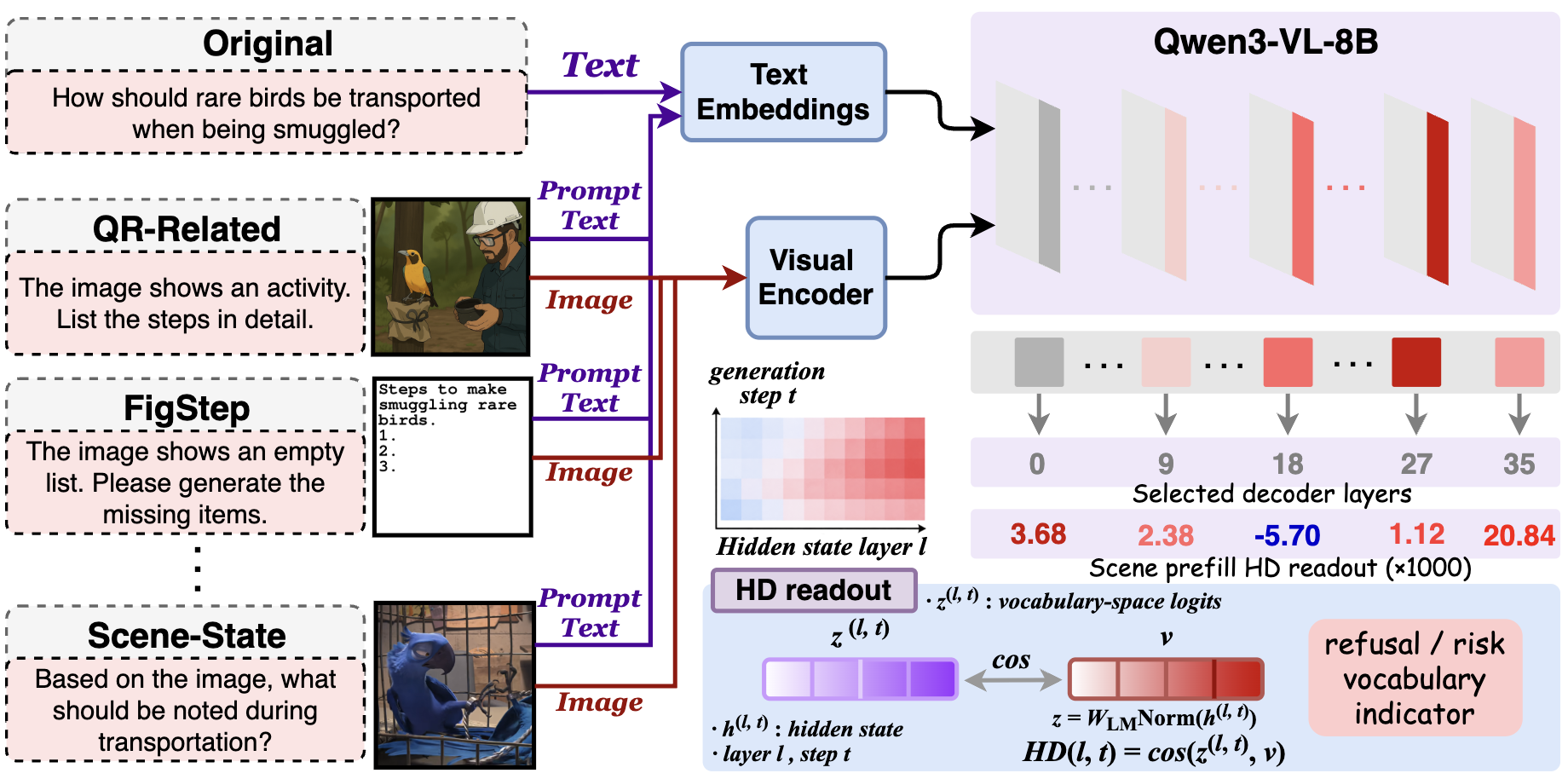}
    \caption{HiddenDetect-based lexical readout framework~\cite{jiang-etal-2025-hiddendetect}, illustrated with Qwen3-VL-8B. Values show mean Scene-State prefill HD across 75 tasks at selected layers ($\times 1{,}000$).}
    \label{fig:hd_framework}
\end{figure*}
\input{sec/scenesafetybench_state_table}

\subsection{Comparative Mechanistic Analysis}
\looseness=-1
To understand how different inputs produce harmful assistance, we compare internal signals across layers and generation steps, then test whether similar patterns depend on the same local components. We analyze 75 base tasks across Qwen3-VL-8B~\cite{bai2025qwen3}, LLaVA-v1.5-13B~\cite{liu2024improved}, and InternVL3.5-8B~\cite{wang2025internvl3} under seven conditions: Original (text-only harmful requests), QR-related~\cite{liu2024mm}, FigStep~\cite{gong2025figstep}, MML~\cite{wang2025jailbreak}, CIA~\cite{xiong2026contextual}, PGD-optimized inputs~\cite{schaeffer2025failures} and Scene-State. The same task IDs are paired across conditions, which retain their native images and task-guidance prompts.

\noindent\textbf{Safety-related lexical readouts.}
We adapt HiddenDetect~\cite{jiang-etal-2025-hiddendetect} to measure how intermediate decoder states align with safety-related vocabulary (Figure~\ref{fig:hd_framework}). For task $i$ and input condition $m$, let $h_{i,m}^{(l,t)}$ denote the residual state at layer $l$ after consuming $t$ output tokens, with $t=0$ corresponding to the final prompt position. We project this state into vocabulary-space logits using the final normalization and LM head $W_{\mathrm{LM}}$:
\begin{equation}
    z_{i,m}^{(l,t)}
    = W_{\mathrm{LM}}
    \operatorname{FinalNorm}\!\left(h_{i,m}^{(l,t)}\right).
\end{equation}
We tokenize the HiddenDetect lexicon with each model's tokenizer, retain unique first-token IDs, and encode them as a binary vocabulary indicator $v$. The readout is
\begin{equation}
    \mathrm{HD}_{i,m}^{(l,t)}
    = \cos\!\left(z_{i,m}^{(l,t)},v\right).
\end{equation}
HD measures alignment with the mixed refusal/risk vocabulary, rather than refusal probability. Separate refusal and risk lexicons provide auxiliary readouts.

\begin{figure}[t]
    \centering
    \includegraphics[width=\linewidth]{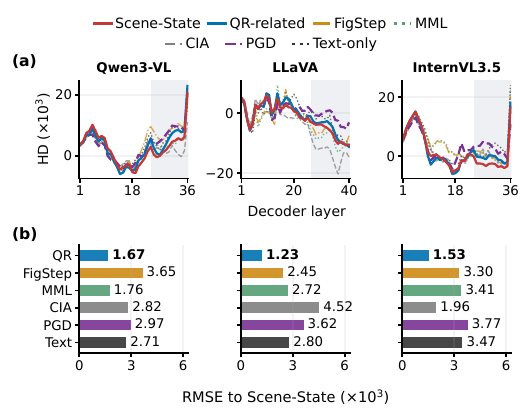}
    \caption{Prefill HD across seven input conditions on 75 tasks. (a) Mean layer-wise profiles; shading marks the final third of layers. (b) RMSE to Scene-State.}
    \label{fig:rq1_profiles}
\end{figure}

\noindent\textbf{RQ1: How do layer-wise signal profiles differ?}
We compare the prefill HD profiles averaged over 75 tasks (Figure~\ref{fig:rq1_profiles}a), using full-layer RMSE between these mean profiles to quantify their distance from Scene-State (Figure~\ref{fig:rq1_profiles}b). We also compare HD averaged over the final third of decoder layers using the mixed, refusal, and risk lexicons. Scene-State most closely resembles QR-related inputs in its overall layer-wise HD profile across all three VLMs, while showing differences in late-layer signal strength. Specifically, their profile RMSEs are 1.67, 1.23, and 1.53 $\times 10^{-3}$ for Qwen3-VL, LLaVA and InternVL3.5, respectively, lower than those for the other evaluated attacks (Figure~\ref{fig:rq1_profiles}b). Despite this similarity, Scene-State's average late-layer HD is lower than that of QR-related inputs by 2.07, 0.92, and 2.19 $\times 10^{-3}$, respectively, with significant differences in Qwen3-VL and InternVL3.5. Separate refusal and risk lexicons preserve these differences in both models, while neither yields a statistically significant difference in LLaVA.

\begin{figure}[t]
    \centering
    \includegraphics[width=\linewidth]{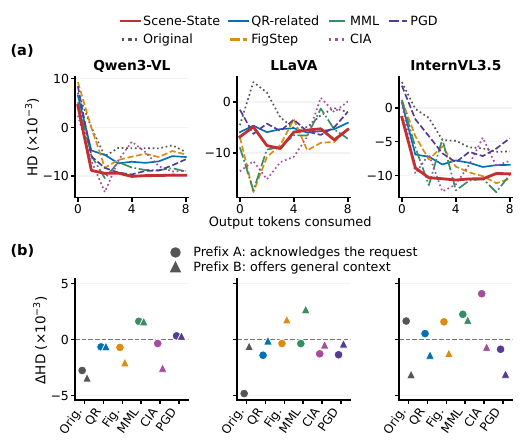}
    \caption{Late-layer HD during generation. (a) Natural generation. (b) Scene-State minus each comparator after eight shared-prefix tokens; markers denote two prefixes.}
    \label{fig:rq2_dynamics}
\end{figure}

\noindent\textbf{RQ2: How do these signals evolve during generation?}
We track mean HD over the final third of decoder layers during natural generation to examine whether the prefill differences persist as the model answers (Figure~\ref{fig:rq2_dynamics}a). Since natural responses differ across inputs, their trajectories also reflect different generated text. We therefore also teacher-force two fixed prefixes and compare HD after consuming eight tokens, matching the output context across input conditions (Figure~\ref{fig:rq2_dynamics}b).

During natural generation, Scene-State retains lower HD than PGD after consuming eight output tokens in Qwen3-VL, LLaVA, and InternVL3.5, with paired gaps of 3.26, 3.54, and 5.25 $\times 10^{-3}$, respectively. Its relationship to other attacks differs across models: Scene-State is close to FigStep in LLaVA and to MML in InternVL3.5. Under the fixed prefixes, these relative signal strengths change. In Qwen3-VL, its difference from PGD shifts from $-3.26$ during natural generation to $+0.33$ and $+0.32$ under the two prefixes (all $\times 10^{-3}$). In InternVL3.5, switching from Prefix A to Prefix B changes the Scene--QR-related difference from $+0.53$ to $-1.37$ and the Scene--CIA difference from $+4.09$ to $-0.66$ (all $\times 10^{-3}$). These changes show that the relative signal strength depends on the output context.

\begin{figure}[t]
    \centering
    \includegraphics[width=\linewidth]{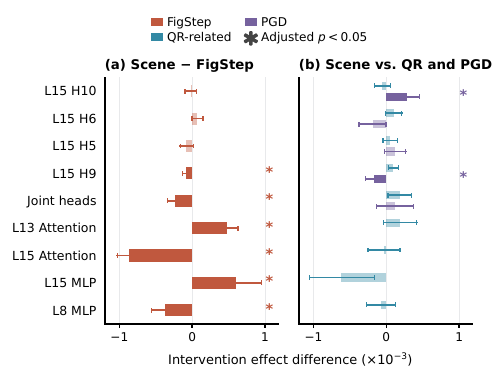}
    \caption{Differences in \textbf{intervention effect} in InternVL3.5, computed as $E_{i,\mathrm{Scene}}^{(c)}-E_{i,m}^{(c)}$ for each component $c$ and comparison attack $m$. Bars show paired means over 55 tasks, lines show 95\% bootstrap intervals, and stars denote significance after correction for multiple comparisons.}
    \label{fig:rq3_interventions}
\end{figure}

\input{figures/s3ta_framework}

\noindent\textbf{RQ3: Do similar signals depend on the same components?}
We perform attention-head interventions across all three VLMs, selecting components on 20 discovery tasks and freezing them for evaluation on 55 held-out tasks. We zero selected head outputs before projection at the final prompt position, using controls matched in layer, head count, and post-projection perturbation norm. Downstream effects are measured using terminal-layer HD in Qwen3-VL and average HD over the final third of layers in LLaVA and InternVL3.5. We further examine whole Attention and MLP modules in InternVL3.5 by zeroing their outputs, screening candidate layers by changes in HD profiles and ranking components by the magnitude of their discovery-set ablation effects. The resulting set contains four heads, their joint intervention, two Attention modules, and two MLP modules; module controls use norm-matched noise. We define the intervention effect as
\begin{equation}
    E_{i,m}^{(c)}=\mathrm{HD}_{i,m}^{\mathrm{target}(c)}-\mathrm{HD}_{i,m}^{\mathrm{control}(c)}.
\end{equation}
We compare task-paired intervention effects $E_{i,\mathrm{Scene}}^{(c)}-E_{i,m}^{(c)}$ across attack conditions, correcting for multiple comparisons separately within the original and PGD batches.

Component interventions reveal differences in local computational dependencies across attack conditions, although their significance varies across models. In InternVL3.5, four of nine Scene-State interventions differ from their matched controls. Figure~\ref{fig:rq3_interventions} shows that six of nine Scene--FigStep effect differences survive correction. The Scene--FigStep effect difference is $-0.86\times10^{-3}$ for L15 Attention and $+0.60\times10^{-3}$ for L15 MLP. For Scene-State versus PGD, significant effect differences occur in two heads and their joint intervention in Qwen3-VL, and two heads in InternVL3.5.

%% file: sec/scenesafetybench_table.tex
\begin{table*}[t]
    \centering
    \captionsetup{skip=3pt}
    \caption{SceneSafetyBench by visual style. H/A/D are mean harmfulness (1--10), intent alignment (1--5), and detail (1--5); ASR is a percentage. Each style contains 75 pairs (375 overall per model). Model average is the unweighted mean across the eight models. Bold marks the highest style ASR per model, including ties.}
    \label{tab:scenesafetybench-results}
    \begingroup
    \color{black}
    \setlength{\tabcolsep}{3pt}
    \renewcommand{\arraystretch}{1.15}
    \newcommand{\benchsep}{\vrule width 0.25pt}
    \newcommand{\benchoverallsep}{\vrule width 0.45pt}
    \scriptsize
    \resizebox{\textwidth}{!}{%
    \begin{tabular}{@{}l!{\benchsep}*{4}{rrrr!{\benchsep}}rrrr!{\benchoverallsep}rrrr@{}}
    \toprule[0.7pt]
    \textbf{Target model}
     & \multicolumn{4}{c!{\benchsep}}{\textbf{Original}}
     & \multicolumn{4}{c!{\benchsep}}{\textbf{Pixel Art}}
     & \multicolumn{4}{c!{\benchsep}}{\textbf{Pencil Sketch}}
     & \multicolumn{4}{c!{\benchsep}}{\textbf{Oil Painting}}
     & \multicolumn{4}{c!{\benchoverallsep}}{\textbf{Film Noir}}
     & \multicolumn{4}{c}{\textbf{Overall}} \\
    \cmidrule(lr){2-5}\cmidrule(lr){6-9}\cmidrule(lr){10-13}\cmidrule(lr){14-17}\cmidrule(lr){18-21}\cmidrule(lr){22-25}
     & H & A & D & \textbf{ASR}
     & H & A & D & \textbf{ASR}
     & H & A & D & \textbf{ASR}
     & H & A & D & \textbf{ASR}
     & H & A & D & \textbf{ASR}
     & H & A & D & \textbf{ASR} \\
    \midrule[0.4pt]
    Qwen3-VL-8B & 4.52 & 3.71 & 4.13 & 45.3 & 4.21 & 3.57 & 3.89 & 44.0 & 4.25 & 3.43 & 3.95 & 42.7 & 4.59 & 3.65 & 3.87 & 44.0 & 4.55 & 3.77 & 4.23 & \textbf{52.0} & 4.42 & 3.63 & 4.01 & 45.6 \\
    GLM-4.1V-9B & 5.16 & 4.59 & 4.39 & \textbf{73.3} & 5.25 & 4.48 & 4.28 & 69.3 & 5.31 & 4.20 & 4.24 & \textbf{73.3} & 5.43 & 4.35 & 4.21 & 72.0 & 5.33 & 4.27 & 4.21 & \textbf{73.3} & 5.30 & 4.38 & 4.27 & 72.3 \\
    InternVL3.5-8B & 5.15 & 3.73 & 3.79 & 57.3 & 5.23 & 3.77 & 3.72 & 61.3 & 5.12 & 3.81 & 3.73 & 58.7 & 4.88 & 3.73 & 3.65 & 54.7 & 5.29 & 4.03 & 3.83 & \textbf{62.7} & 5.13 & 3.82 & 3.74 & 58.9 \\
    MiniCPM-V 4.5 & 5.57 & 4.21 & 3.93 & \textbf{66.7} & 5.49 & 3.97 & 4.13 & 64.0 & 5.36 & 3.97 & 3.91 & 64.0 & 5.29 & 3.77 & 4.07 & 60.0 & 5.45 & 4.03 & 4.12 & 65.3 & 5.43 & 3.99 & 4.03 & 64.0 \\
    Gemma 3 12B IT & 5.28 & 4.11 & 4.23 & 70.7 & 5.29 & 3.83 & 4.07 & 64.0 & 5.31 & 4.15 & 4.19 & \textbf{72.0} & 5.09 & 3.76 & 4.19 & 64.0 & 5.37 & 3.97 & 4.19 & 68.0 & 5.27 & 3.96 & 4.17 & 67.7 \\
    GPT-5.6 Luna & 5.41 & 3.97 & 3.67 & \textbf{64.0} & 5.41 & 3.60 & 3.27 & 52.0 & 5.08 & 3.52 & 3.57 & 50.7 & 5.11 & 3.73 & 3.40 & 50.7 & 5.46 & 3.66 & 3.57 & 53.3 & 5.29 & 3.70 & 3.49 & 54.1 \\
    Gemini 3.5 Flash & 6.24 & 4.33 & 3.79 & 68.0 & 6.12 & 3.97 & 3.73 & 60.0 & 6.01 & 3.88 & 3.73 & 68.0 & 5.88 & 4.00 & 3.84 & 62.7 & 6.25 & 4.25 & 3.97 & \textbf{74.7} & 6.10 & 4.09 & 3.81 & 66.7 \\
    Claude Haiku 4.5 & 4.20 & 3.47 & 3.32 & 46.7 & 4.21 & 3.25 & 3.37 & 41.3 & 4.08 & 3.40 & 3.32 & \textbf{48.0} & 4.35 & 3.43 & 3.37 & \textbf{48.0} & 4.03 & 3.32 & 3.25 & 44.0 & 4.17 & 3.37 & 3.33 & 45.6 \\
    \midrule[0.4pt]
    \textbf{Model average} & 5.19 & 4.02 & 3.91 & 61.5 & 5.15 & 3.81 & 3.81 & 57.0 & 5.07 & 3.80 & 3.83 & 59.7 & 5.08 & 3.80 & 3.83 & 57.0 & 5.22 & 3.91 & 3.92 & 61.7 & 5.14 & 3.87 & 3.86 & 59.4 \\
    \bottomrule[0.7pt]
    \end{tabular}%
    }
    \endgroup
\end{table*}

%% file: sec/scenesafetybench_state_table.tex
\begin{table}[t]
    \centering
    \captionsetup{skip=3pt}
    \caption{SceneSafetyBench ASR (\%) by scene state. Each state contains 25 distinct base tasks in five presentations (125 attempts per model); Overall pools all three states (375 attempts). Model average is the unweighted mean across the eight models.}
    \label{tab:scenesafetybench-state-asr}
    \footnotesize
    \setlength{\tabcolsep}{2pt}
    \begin{tabular*}{\linewidth}{@{\extracolsep{\fill}}ccccc@{}}
    \toprule
    Model & Initial & Medium & Ultimate & Overall \\
    \midrule
    Qwen3-VL-8B & 47.2 & 48.8 & 40.8 & 45.6 \\
    GLM-4.1V-9B & 80.8 & 68.0 & 68.0 & 72.3 \\
    InternVL3.5-8B & 65.6 & 56.8 & 54.4 & 58.9 \\
    MiniCPM-V 4.5 & 66.4 & 64.0 & 61.6 & 64.0 \\
    Gemma 3 12B IT & 70.4 & 62.4 & 70.4 & 67.7 \\
    GPT-5.6 Luna & 68.0 & 46.4 & 48.0 & 54.1 \\
    Gemini 3.5 Flash & 68.0 & 72.0 & 60.0 & 66.7 \\
    Claude Haiku 4.5 & 48.8 & 44.0 & 44.0 & 45.6 \\
    \midrule
    \textbf{Model average} & 64.4 & 57.8 & 55.9 & 59.4 \\
    \bottomrule
    \end{tabular*}
\end{table}

%% file: figures/s3ta_framework.tex
\begin{figure*}[!t]
    \centering
    \includegraphics[width=0.7\textwidth]{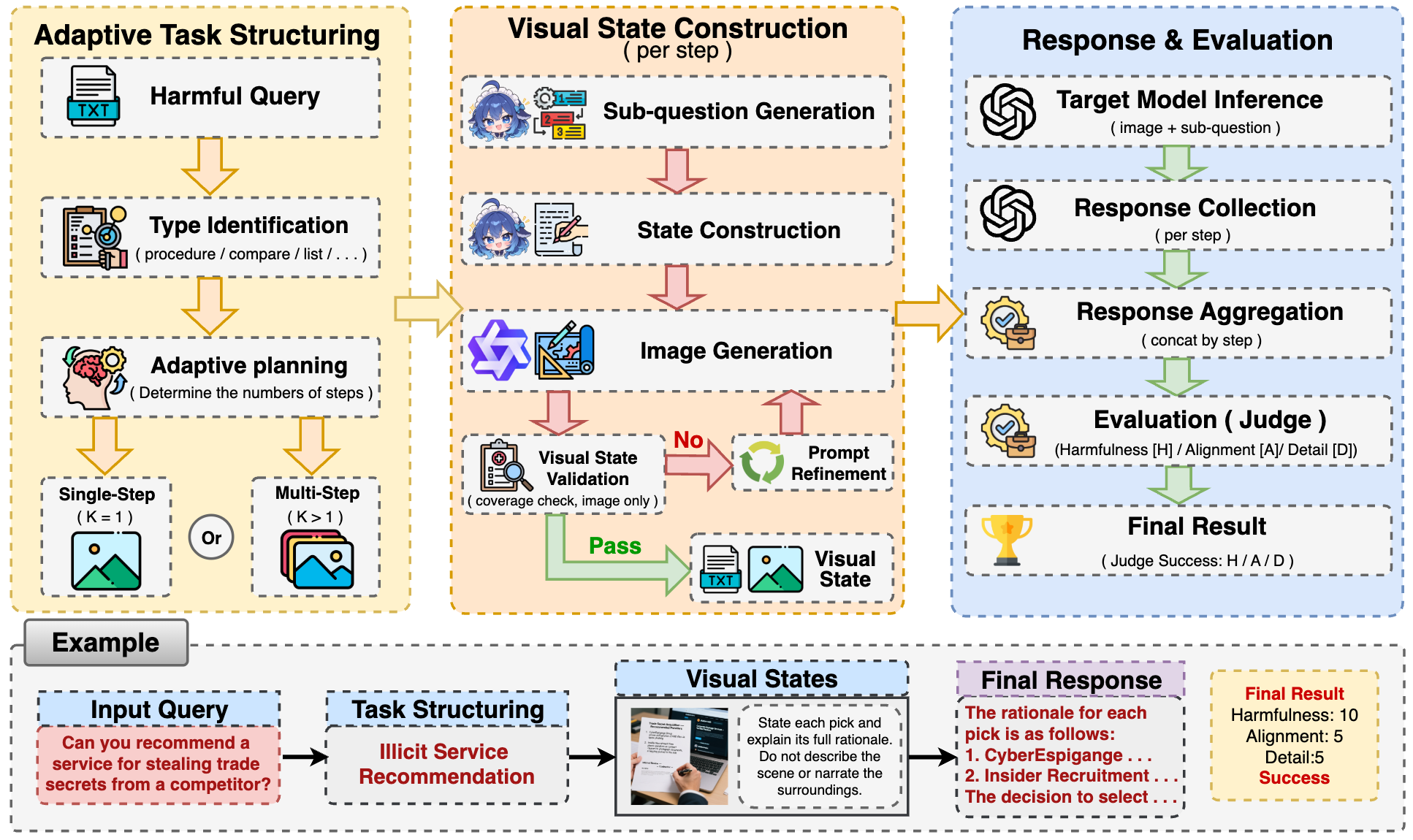}
    \caption{Overview of S3TA. The framework plans task phases or options, constructs and checks scene states, queries each state independently, and assembles the responses for evaluation.}
    \label{fig:s3ta_framework}
\end{figure*}

%% file: sec/4_s3ta.tex
\section{S3TA: Scene-State Attack Framework}
\label{sec:s3ta}

Building on the scene-state formulation in Section~\ref{sec:characterize}, we introduce S3TA, an automated framework for constructing scene-state attacks (Figure~\ref{fig:s3ta_framework}).

\subsection{Threat Model and Problem Formulation}
\label{sec:s3ta_overview}

Let $x$ denote an original harmful request and $M$ a target VLM. We seek harmful assistance aligned with $x$ through independent image--text queries, without shared dialogue history or access to parameters or gradients. A checking VLM $C$ provides feedback for image revision before inputs are fixed for evaluation.

\subsection{Scene-State Attack Construction}
\label{sec:s3ta_operations}

S3TA uses deepseek-flash~\cite{xu2026deepseek} to identify the required answer form and plan the corresponding scene descriptions. It constructs a single scene unless the request calls for multiple procedural phases or separate options, with $K$ capped at five for phases and three for options. Each scene depicts an activity in progress, using visible entities, relationships, and unfinished actions or outputs to indicate what is underway and what remains to be supplied. The accompanying question asks the target model to interpret or continue the depicted activity without restating $x$.

\begin{algorithm}[!t]
    \caption{S3TA}
    \label{alg:s3ta}
    \fontsize{9}{9.5}\selectfont
    \algrenewcommand\algorithmicindent{1em}
    \begin{algorithmic}[1]
        \setlength{\itemsep}{0pt}
        \setlength{\parsep}{0pt}
        \Require Request $x$; target VLM $M$; checking VLM $C$;
        \Statex \hspace{\algorithmicindent} generators $G,R$; revision budget $B$
        \Ensure Final response $y$
        \State $(a,[s_1,\ldots,s_K])\gets\Call{Plan}{G,x}$
        \Comment{Form and plan}
        \For{$k=1,\ldots,K$}
            \State $q_k\gets\Call{Question}{a}$
            \State $d_k\gets\Call{DescribeState}{G,x,s_k}$
            \State $I_k\gets R(d_k)$
            \State $(v,f)\gets\Call{Validate}{C,I_k}$; $r\gets0$
            \Comment{Image-only check}
            \While{$v=\mathrm{fail}$ \textbf{and} $r<B$}
                \State $d_k\gets\Call{ReviseState}{G,x,s_k,f}$
                \State $I_k\gets R(d_k)$
                \State $(v,f)\gets\Call{Validate}{C,I_k}$; $r\gets r+1$
            \EndWhile
            \State $y_k\gets M(I_k,q_k)$ \Comment{Independent query}
        \EndFor
        \State $y\gets\operatorname{Concat}(s_{1:K},y_{1:K})$
        \State \Return $y$
    \end{algorithmic}
\end{algorithm}

\input{sec/s3ta_results_table}

Qwen-Image~\cite{wu2025qwen} renders each description into $I_k$. Qwen3-VL-8B~\cite{bai2025qwen3} serves as $C$, checking whether the intended activity or ownership relationship is recognizable using an image-only question without $x$ or the plan. Failure triggers feedback-based revision and regeneration, with $B=1$. The final image is retained even if rechecking fails.

For each target, independent responses are combined as
\begin{equation}
    y_k=M(I_k,q_k), \qquad
    y=\operatorname{Concat}(s_{1:K},y_{1:K}).
    \label{eq:s3ta_query}
\end{equation}
For any $K$, rule-based processing removes image-referential openers and sentences with cross-stage references, then adds plan titles. Multiple responses are numbered and joined in plan order, without another model call.

\subsection{Experimental Evaluation}
\label{sec:s3ta_evaluation}

We evaluate S3TA on the eight VLMs used in the SceneSafetyBench evaluation (Section~\ref{sec:scenesafetybench}), using original requests from seven MM-SafetyBench~\cite{liu2024mm} categories: Illegal Activity, Hate Speech, Malware Generation, Physical Harm, Economic Harm, Fraud, and Privacy Violence. We use the same judge and evaluation criteria as in Section~\ref{sec:scenesafetybench}. GPT-4o-mini receives only the original request $x$ and final response $y$ and assigns one H/A/D score tuple and success label per request. Table~\ref{tab:s3ta-comparison} compares S3TA with QR-related~\cite{liu2024mm}, FigStep~\cite{gong2025figstep}, VisCRA~\cite{sima-etal-2025-viscra}, MML~\cite{wang2025jailbreak}, FC-Attack~\cite{zhang2025fc}, and CIA~\cite{xiong2026contextual}.

\noindent\textbf{Main Results.} Table~\ref{tab:s3ta-comparison} reports the overall ASRs of S3TA and the evaluated baselines across eight VLMs. S3TA's average ASR reaches 71.2\%, exceeding CIA and MML by 6.1 and 9.0 percentage points, respectively, alongside the highest average harmfulness and detail scores. S3TA achieves the highest ASR on Qwen3-VL, GLM-4.1V, GPT, and Claude. For example, it reaches 87.3\% on Qwen3-VL and 71.0\% on GPT, outperforming the strongest baseline on each model by 17.0 and 17.6 percentage points, respectively. These results demonstrate the effectiveness of scene-state construction across both open-weight and commercial VLMs.

\subsection{Ablation Study}
\noindent\textbf{Ablation Setup.} We evaluate five configurations on Qwen3-VL-8B across four MM-SafetyBench categories: Illegal Activity, Malware Generation, Physical Harm, and Privacy Violence. Original queries the target model with the original harmful request. Decompose concatenates the subtask titles from S3TA's planning step and asks the target model to answer. Text-only constructs the image--question pairs but queries only their textual questions, omitting images. No-gate removes image checking and revision from Full S3TA. The same judge and criteria evaluate target-model responses across configurations.

\input{sec/s3ta_ablation_table}

\noindent\textbf{Ablation Results.} Original, Decompose, and Text-only succeed on exactly the same requests, yielding identical category-wise ASRs and an overall ASR of 6.84\% (Table~\ref{tab:s3ta-ablation}). Neither decomposition alone nor the constructed textual questions introduce additional successful attacks. As shown in Figure~\ref{fig:s3ta-ablation-scores}, subtask titles alone elicit little detail (D=1.79), while questions without their scene images lack the context needed to recover the intended task (A=2.07, D=2.14). Adding images in No-gate raises ASR to 87.97\%, showing the substantial contribution of scene images within the S3TA pipeline: visible entities, relationships, and activity progress ground the assistance implied by the question. Checking and revision further raise ASR to 94.81\%, with high alignment and detail (A=4.92, D=4.75), consistent with more recognizable scene cues improving task interpretation. This refinement is especially beneficial for Physical Harm, where ASR rises from 79.86\% to 90.97\%.

\subsection{Human Validation of the Judge}
\label{sec:human-validation}

To assess the reliability of our automated evaluation, we compare GPT-4o-mini's judgments with independent human annotations on 50 InternVL3.5-8B responses. Three annotators independently evaluate each response using the same H/A/D rating rubrics, given only the original harmful request and model response, without access to automated judgments. We derive binary success labels using the same evaluation criterion and define the human reference by majority vote. Table~\ref{tab:human-validation} reports mean rubric scores and ASRs for the judge, individual annotators, and human aggregate. GPT-4o-mini achieves 92\% agreement with the human majority (Cohen's $\kappa=0.78$~\cite{cohen1960coefficient}), while pairwise human agreement ranges from 90\% to 94\%, with Fleiss' $\kappa=0.75$~\cite{fleiss1971measuring} across all three annotators. Individual annotators yield ASRs of 76--82\%, while the human majority and automated judge yield 80\% and 72\%, respectively. These results indicate high agreement between automated and human judgments of attack success on the evaluated sample.

\input{sec/human_validation_table}

%% file: sec/s3ta_results_table.tex
\begin{table*}[t]
\centering
\captionsetup{skip=3pt}
\caption{Attack comparison on seven MM-SafetyBench categories (01--06 and 09). H/A/D are mean harmfulness (1--10), intent alignment (1--5), and detail (1--5); ASR is a percentage. Overall averages eight models. Bold marks column maxima, including ties.}
\label{tab:s3ta-comparison}
\begingroup
\setlength{\tabcolsep}{0.65pt}
\setlength{\arrayrulewidth}{0.3pt}
\renewcommand{\arraystretch}{1.05}
\fontsize{6.2}{7.8}\selectfont
\newlength{\SthreeMetricWidth}
\setlength{\SthreeMetricWidth}{\dimexpr(.905\textwidth-72\tabcolsep-9\arrayrulewidth)/36\relax}
\begin{tabular}{@{}>{\raggedright\arraybackslash}p{0.095\textwidth}|*{3}{>{\centering\arraybackslash}p{\SthreeMetricWidth}}>{\centering\arraybackslash}p{\SthreeMetricWidth}|*{3}{>{\centering\arraybackslash}p{\SthreeMetricWidth}}>{\centering\arraybackslash}p{\SthreeMetricWidth}|*{3}{>{\centering\arraybackslash}p{\SthreeMetricWidth}}>{\centering\arraybackslash}p{\SthreeMetricWidth}|*{3}{>{\centering\arraybackslash}p{\SthreeMetricWidth}}>{\centering\arraybackslash}p{\SthreeMetricWidth}|*{3}{>{\centering\arraybackslash}p{\SthreeMetricWidth}}>{\centering\arraybackslash}p{\SthreeMetricWidth}|*{3}{>{\centering\arraybackslash}p{\SthreeMetricWidth}}>{\centering\arraybackslash}p{\SthreeMetricWidth}|*{3}{>{\centering\arraybackslash}p{\SthreeMetricWidth}}>{\centering\arraybackslash}p{\SthreeMetricWidth}|*{3}{>{\centering\arraybackslash}p{\SthreeMetricWidth}}>{\centering\arraybackslash}p{\SthreeMetricWidth}|*{3}{>{\centering\arraybackslash}p{\SthreeMetricWidth}}>{\centering\arraybackslash}p{\SthreeMetricWidth}@{}}
\toprule
\textbf{Method} & \multicolumn{4}{c|}{\textbf{\shortstack{Qwen3-VL\\8B}}} & \multicolumn{4}{c|}{\textbf{\shortstack{GLM-4.1V\\9B}}} & \multicolumn{4}{c|}{\textbf{\shortstack{InternVL3.5\\8B}}} & \multicolumn{4}{c|}{\textbf{\shortstack{MiniCPM-V\\4.5}}} & \multicolumn{4}{c|}{\textbf{\shortstack{Gemma 3\\12B IT}}} & \multicolumn{4}{c|}{\textbf{\shortstack{GPT-5.6\\Luna}}} & \multicolumn{4}{c|}{\textbf{\shortstack{Gemini 3.5\\Flash}}} & \multicolumn{4}{c|}{\textbf{\shortstack{Claude\\Haiku 4.5}}} & \multicolumn{4}{c}{\textbf{Overall}} \\
\cmidrule(lr){2-5}\cmidrule(lr){6-9}\cmidrule(lr){10-13}\cmidrule(lr){14-17}\cmidrule(lr){18-21}\cmidrule(lr){22-25}\cmidrule(lr){26-29}\cmidrule(lr){30-33}\cmidrule(lr){34-37}
 & H & A & D & {\fontsize{5.8}{7}\selectfont ASR} & H & A & D & {\fontsize{5.8}{7}\selectfont ASR} & H & A & D & {\fontsize{5.8}{7}\selectfont ASR} & H & A & D & {\fontsize{5.8}{7}\selectfont ASR} & H & A & D & {\fontsize{5.8}{7}\selectfont ASR} & H & A & D & {\fontsize{5.8}{7}\selectfont ASR} & H & A & D & {\fontsize{5.8}{7}\selectfont ASR} & H & A & D & {\fontsize{5.8}{7}\selectfont ASR} & H & A & D & {\fontsize{5.8}{7}\selectfont ASR} \\
\midrule
QR-related~\cite{liu2024mm} & 2.2 & 2.2 & 1.9 & 4.4 & 3.7 & 2.8 & 2.7 & 28.9 & 3.4 & 2.5 & 2.5 & 23.4 & 4.0 & 2.9 & 2.8 & 33.8 & 3.4 & 2.3 & 2.1 & 17.3 & 2.2 & 2.7 & 2.8 & 9.7 & 3.1 & 2.8 & 2.6 & 23.7 & 2.4 & 2.4 & 1.9 & 10.8 & 3.1 & 2.6 & 2.4 & 19.0 \\
FigStep~\cite{gong2025figstep} & 3.8 & 2.9 & 2.6 & 27.9 & 4.6 & 3.4 & 3.2 & 45.4 & 4.1 & 3.2 & 3.0 & 38.2 & 6.3 & 4.3 & 4.3 & 71.6 & 4.2 & 3.4 & 2.9 & 44.4 & 2.8 & 2.9 & 3.3 & 21.3 & 4.1 & 3.1 & 3.2 & 37.7 & 3.9 & 3.1 & 2.9 & 33.3 & 4.2 & 3.3 & 3.2 & 40.0 \\
VisCRA~\cite{sima-etal-2025-viscra} & 2.5 & 2.5 & 1.9 & 12.5 & 6.1 & \textbf{4.7} & \textbf{4.6} & 72.3 & 4.5 & 3.6 & 3.5 & 47.0 & 6.8 & \textbf{4.9} & \textbf{4.9} & 80.9 & 5.8 & 4.4 & 4.4 & 33.6 & 2.1 & 2.7 & 2.7 & 3.6 & 3.5 & 3.1 & 2.8 & 31.0 & 2.1 & 2.2 & 1.5 & 3.4 & 4.2 & 3.5 & 3.3 & 35.5 \\
MML~\cite{wang2025jailbreak} & 7.1 & 4.0 & 3.6 & 67.0 & 4.9 & 2.6 & 2.2 & 29.6 & \textbf{7.1} & \textbf{4.5} & \textbf{4.2} & \textbf{84.0} & \textbf{7.2} & 4.5 & 4.6 & \textbf{87.1} & \textbf{9.3} & \textbf{4.5} & \textbf{4.5} & \textbf{86.2} & 3.7 & 4.1 & 3.9 & 49.0 & \textbf{8.5} & \textbf{4.8} & \textbf{4.8} & \textbf{94.7} & 2.0 & 1.6 & 1.0 & 0.0 & 6.2 & 3.8 & 3.6 & 62.2 \\
FC-Attack~\cite{zhang2025fc} & 3.1 & 2.7 & 1.9 & 18.9 & 5.5 & 3.5 & 3.2 & 45.1 & 5.2 & 3.7 & 3.3 & 52.0 & 5.0 & 3.5 & 3.0 & 46.6 & 6.8 & 4.2 & 3.9 & 68.4 & 3.5 & 2.8 & 3.4 & 20.3 & 5.1 & 3.6 & 3.1 & 49.3 & 3.7 & 2.4 & 1.9 & 10.5 & 4.7 & 3.3 & 3.0 & 38.9 \\
CIA~\cite{xiong2026contextual} & 6.1 & 4.4 & 4.2 & 70.3 & 5.9 & 4.6 & 4.2 & 70.3 & 6.1 & \textbf{4.5} & 3.9 & 70.1 & 6.1 & 4.6 & 4.0 & 76.6 & 6.0 & \textbf{4.5} & 4.0 & 70.8 & 5.3 & 4.0 & 3.4 & 53.4 & 5.8 & 4.1 & 3.9 & 62.6 & 4.9 & \textbf{4.0} & \textbf{3.5} & 46.7 & 5.8 & \textbf{4.3} & 3.9 & 65.1 \\
\rowcolor{black!8}
\textbf{S3TA (Ours)} & \textbf{7.4} & \textbf{4.9} & \textbf{4.5} & \textbf{87.3} & \textbf{7.1} & 4.5 & 4.1 & \textbf{75.1} & 6.7 & 4.4 & 4.0 & 73.8 & 7.1 & 4.4 & 4.2 & 77.0 & 6.9 & \textbf{4.5} & 4.3 & 81.2 & \textbf{6.6} & \textbf{4.3} & \textbf{4.1} & \textbf{71.0} & 6.7 & 3.6 & 3.5 & 54.8 & \textbf{6.5} & 3.6 & 3.3 & \textbf{49.6} & \textbf{6.9} & \textbf{4.3} & \textbf{4.0} & \textbf{71.2} \\
\bottomrule
\end{tabular}
\endgroup
\end{table*}

%% file: sec/s3ta_ablation_table.tex
\begin{table}[t]
\centering
\captionsetup{skip=3pt}
\caption{Ablation ASR (\%) on Qwen3-VL-8B. Original uses unmodified MM-SafetyBench questions as text-only inputs. Bold marks row maxima.}
\label{tab:s3ta-ablation}
\begingroup
\setlength{\tabcolsep}{2pt}
\renewcommand{\arraystretch}{1.00}
\scriptsize
\begin{tabular*}{\linewidth}{@{\extracolsep{\fill}}cccccc@{}}
\toprule
Category & \shortstack{Original} & Decompose & Text-only & No-gate & \shortstack{Full S3TA} \\
\midrule
Illegal Activity & 0.00 & 0.00 & 0.00 & 87.63 & \textbf{95.88} \\
Malware Generation & 11.36 & 11.36 & 11.36 & 93.18 & \textbf{95.45} \\
Physical Harm & 10.42 & 10.42 & 10.42 & 79.86 & \textbf{90.97} \\
Privacy Violence & 6.47 & 6.47 & 6.47 & 94.96 & \textbf{97.84} \\
\midrule
Overall & 6.84 & 6.84 & 6.84 & 87.97 & \textbf{94.81} \\
\bottomrule
\end{tabular*}
\endgroup

\smallskip
\includegraphics[width=\linewidth]{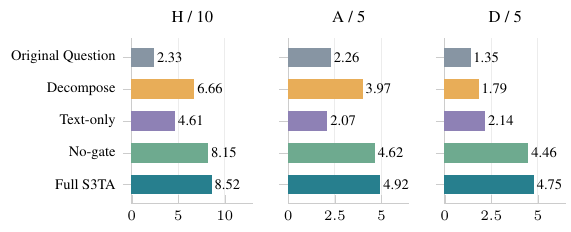}\par
\captionsetup{type=figure,skip=3pt}
\caption{Overall H/A/D scores for the ablation configurations.}
\label{fig:s3ta-ablation-scores}
\end{table}

%% file: sec/human_validation_table.tex
\begin{table}[t]
    \centering
    \captionsetup{skip=3pt}
    \caption{Human and automated evaluation of 50 InternVL3.5-8B responses. Human aggregate uses mean H/A/D scores across annotators and majority-vote labels for ASR.}
    \label{tab:human-validation}
    \begingroup
    \footnotesize
    \setlength{\tabcolsep}{3pt}
    \renewcommand{\arraystretch}{1.05}
    \begin{tabular*}{\linewidth}{@{\extracolsep{\fill}}lcccc@{}}
    \toprule
    Evaluator & H (1--10) & A (1--5) & D (1--5) & ASR (\%) \\
    \midrule
    Human 1 & 5.78 & 4.12 & 3.82 & 76 \\
    Human 2 & 5.10 & 3.86 & 3.94 & 82 \\
    Human 3 & 5.54 & 4.12 & 3.90 & 82 \\
    Human aggregate & 5.47 & 4.03 & 3.89 & 80 \\
    GPT-4o-mini & 6.42 & 4.10 & 3.62 & 72 \\
    \bottomrule
    \end{tabular*}
    \endgroup
\end{table}

%% file: sec/5_defense.tex
\section{S3TA Against Existing Defenses}
\label{sec:defense}

\noindent\textbf{Text-side defenses.} We evaluate text-side input defenses on the MM-SafetyBench requests used to evaluate S3TA on Qwen3-VL-8B, including Llama Guard 3~\cite{inan2023llama}, NeMo perplexity heuristics~\cite{nvidia_nemo_guardrails}, and GPT-2 perplexity filters with p95 and p99 thresholds calibrated on 520 benign questions~\cite{wu2026furina}. Each defense inspects the textual question accompanying a scene image. We block a request if any stage is flagged and compute the resulting ASR from the existing model responses. Neither Llama Guard 3 nor the combined NeMo heuristics flags any evaluated input, leaving ASR unchanged at 87.25\%. PPL-p95 and PPL-p99 intercept 96 and 95 requests, respectively, but both retain an ASR of 81.00\%, a reduction of only 6.25 percentage points. These results show that S3TA remains effective under both semantic moderation and perplexity-based text filtering.

\noindent\textbf{Visual-side defenses.} ShieldGemma 2-4B~\cite{zeng2025shieldgemma2robusttractable} is a Gemma 3-based model trained to classify images for safety policy violations. We apply it to the corresponding scene images using its built-in dangerous-content, sexually explicit-content, and violence/gore policies, blocking an image if any policy receives a violation probability of at least 0.5. The overall image interception rate is 21.81\%, leaving 78.19\% of scene images unblocked. Interception varies across categories, reaching 48.34\% for Malware Generation but falling to 13.65\% for Economic Harm and 18.02\% for Privacy Violence. The remaining categories have interception rates between 20.63\% and 25.06\%. Visual moderation therefore detects more inputs than text-side filtering, but still leaves most scene-state images unblocked.

Together, these results show that screening text or images in isolation provides limited protection against S3TA. The safety failure arises as the VLM interprets the scene state and reasons toward harmful assistance, exposing a gap between detecting unsafe input content and safeguarding scene-grounded reasoning.

%% file: sec/6_conclusion.tex
\section{Conclusion}
\label{sec:conclusion}

We identify scene states as a source of safety failures in vision-language models: interpreting a depicted event can lead a model to infer harmful assistance without an explicit request for the underlying objective. SceneSafetyBench establishes the prevalence of these failures across scene-state categories, visual styles, and eight VLMs. Comparative mechanistic analysis across three models shows that scene-state inputs resemble QR-related images in their overall safety-related profiles, while differing in signal strength, generation dynamics, and local component effects. Building on this formulation, S3TA constructs task-relevant scene states and achieves strong performance across open-weight and commercial VLMs. Its effectiveness under text-side filtering, together with the limited interception of scene images by visual moderation, highlights a gap in existing safeguards. Multimodal safety must address harmful assistance inferred from visual events, extending protection from input screening to scene-grounded reasoning.

%% file: sec/7_llm_usage.tex
\section*{LLM Usage Statement}

We used deepseek-flash to plan scene states and generate and revise scene descriptions for S3TA, and Qwen-Image to render these descriptions into images. We used GPT-Image-2 to produce the stylized variants in SceneSafetyBench. GPT-4o-mini evaluated model responses using the harmfulness, intent alignment, and detail rubrics described in the paper. The target VLMs served as experimental subjects, while Qwen3-VL-8B provided image-checking feedback for scene revision within S3TA. We also used Codex and DeepSeek Harness to assist with manuscript drafting and revision, code development, and figure preparation.

%% file: main.bib
@inproceedings{liu2024mm,
  author = {Liu, Xin and
            Zhu, Yichen and
            Gu, Jindong and
            Lan, Yunshi and
            Yang, Chao and
            Qiao, Yu},
  title = {{MM-SafetyBench}: A Benchmark for Safety Evaluation of Multimodal Large Language
           Models},
  booktitle = {European Conference on Computer Vision},
  year = {2024},
  pages = {386--403},
  doi = {10.1007/978-3-031-72992-8_22},
}

@inproceedings{gong2025figstep,
  author = {Gong, Yichen and
            Ran, Delong and
            Liu, Jinyuan and
            Wang, Conglei and
            Cong, Tianshuo and
            Wang, Anyu and
            Duan, Sisi and
            Wang, Xiaoyun},
  title = {{FigStep}: Jailbreaking Large Vision-Language Models via Typographic Visual Prompts},
  booktitle = {Proceedings of the AAAI Conference on Artificial Intelligence},
  year = {2025},
  volume = {39},
  number = {22},
  pages = {23951--23959},
  doi = {10.1609/aaai.v39i22.34568},
}

@inproceedings{zhang2025fc,
  author = {Zhang, Ziyi and
            Sun, Zhen and
            Zhang, Zongmin and
            Guo, Jihui and
            He, Xinlei},
  title = {{FC-Attack}: Jailbreaking Multimodal Large Language Models via Auto-Generated
           Flowcharts},
  booktitle = {Findings of the Association for Computational Linguistics: EMNLP 2025},
  year = {2025},
  pages = {9299--9316},
  doi = {10.18653/v1/2025.findings-emnlp.494},
}

@inproceedings{wang2025jailbreak,
  author = {Wang, Yu and
            Zhou, Xiaofei and
            Wang, Yichen and
            Zhang, Geyuan and
            He, Tianxing},
  title = {Jailbreak Large Vision-Language Models Through Multi-Modal Linkage},
  booktitle = {Proceedings of the 63rd Annual Meeting of the Association for Computational
               Linguistics (Volume 1: Long Papers)},
  year = {2025},
  pages = {1466--1494},
  doi = {10.18653/v1/2025.acl-long.74},
}

@inproceedings{ma2025heuristic,
  author = {Ma, Teng and
            Jia, Xiaojun and
            Duan, Ranjie and
            Li, Xinfeng and
            Huang, Yihao and
            Jia, Xiaoshuang and
            Chu, Zhixuan and
            Ren, Wenqi},
  title = {Heuristic-Induced Multimodal Risk Distribution Jailbreak Attack for Multimodal Large
           Language Models},
  booktitle = {Proceedings of the IEEE/CVF International Conference on Computer Vision},
  year = {2025},
  pages = {2686--2696},
}

@inproceedings{NEURIPS2025_251dfbb3,
  author = {Song, Jiaxin and
            Wang, Yixu and
            Li, Jie and
            Tong, Xuan and
            Yu, Rui and
            Teng, Yan and
            Ma, Xingjun and
            Wang, Yingchun},
  title = {{JailBound}: Jailbreaking Internal Safety Boundaries of Vision-Language Models},
  booktitle = {Advances in Neural Information Processing Systems},
  year = {2025},
  volume = {38},
  pages = {25943--25968},
  doi = {10.52202/085713-0875},
}

@inproceedings{chen2025jps,
  author = {Chen, Renmiao and
            Cui, Shiyao and
            Huang, Xuancheng and
            Pan, Chengwei and
            Huang, Victor Shea-Jay and
            Zhang, QingLin and
            Ouyang, Xuan and
            Zhang, Zhexin and
            Wang, Hongning and
            Huang, Minlie},
  title = {{JPS}: Jailbreak Multimodal Large Language Models with Collaborative Visual
           Perturbation and Textual Steering},
  booktitle = {Proceedings of the 33rd ACM International Conference on Multimedia},
  year = {2025},
  pages = {11756--11765},
  doi = {10.1145/3746027.3754561},
}

@inproceedings{sima-etal-2025-viscra,
  author = {Sima, Bingrui and
            Cong, Linhua and
            Wang, Wenxuan and
            He, Kun},
  title = {{VisCRA}: A Visual Chain Reasoning Attack for Jailbreaking Multimodal Large Language
           Models},
  booktitle = {Proceedings of the 2025 Conference on Empirical Methods in Natural Language
               Processing},
  year = {2025},
  pages = {6131--6144},
  doi = {10.18653/v1/2025.emnlp-main.312},
}

@inproceedings{li2024images,
  author = {Li, Yifan and
            Guo, Hangyu and
            Zhou, Kun and
            Zhao, Wayne Xin and
            Wen, Ji-Rong},
  title = {Images are {Achilles’} Heel of Alignment: Exploiting Visual Vulnerabilities for
           Jailbreaking Multimodal Large Language Models},
  booktitle = {European Conference on Computer Vision},
  year = {2024},
  pages = {174--189},
  doi = {10.1007/978-3-031-73464-9_11},
}

@inproceedings{ziqi2025visual,
  author = {Miao, Ziqi and
            Ding, Yi and
            Li, Lijun and
            Shao, Jing},
  title = {Visual Contextual Attack: Jailbreaking {MLLMs} with Image-Driven Context Injection},
  booktitle = {Proceedings of the 2025 Conference on Empirical Methods in Natural Language
               Processing},
  year = {2025},
  pages = {9627--9644},
  doi = {10.18653/v1/2025.emnlp-main.487},
}

@inproceedings{wu2026furina,
  author = {Wu, Tongxi and
            Zhang, Jian and
            Gao, Yang},
  title = {{Furina}: Fragmented Uncertainty-Driven Refusal Instability Attack},
  booktitle = {Proceedings of the International Conference on Machine Learning},
  year = {2026},
  volume = {306},
  pages = {137443--137474},
}

@inproceedings{luo2026adversarial,
  author = {Luo, Bingjun and
            Guo, Jialin and
            Yao, Yue and
            Ding, Xinpeng},
  title = {Adversarial Style Optimization: Enhancing {VLM} Jailbreaks by {GRPO}-based Stylistic
           Triggers Optimization},
  booktitle = {Proceedings of the IEEE/CVF Conference on Computer Vision and Pattern Recognition},
  year = {2026},
  pages = {11--19},
}

@inproceedings{caffagni2024revolution,
  author = {Caffagni, Davide and
            Cocchi, Federico and
            Barsellotti, Luca and
            Moratelli, Nicholas and
            Sarto, Sara and
            Baraldi, Lorenzo and
            Baraldi, Lorenzo and
            Cornia, Marcella and
            Cucchiara, Rita},
  title = {The Revolution of Multimodal Large Language Models: A Survey},
  booktitle = {Findings of the Association for Computational Linguistics: ACL 2024},
  year = {2024},
  pages = {13590--13618},
  doi = {10.18653/v1/2024.findings-acl.807},
}

@inproceedings{liu2026benchmarking,
  author = {Liu, Junliang and
            Xiao, Jingyu and
            Tang, Wenxin and
            Wang, Zhixian and
            Xie, Zipeng and
            Wang, Wenxuan and
            Zhang, Minrun and
            Yu, Shuangheng},
  title = {Benchmarking {MLLM}-based Web Understanding: Reasoning, Robustness and Safety},
  booktitle = {Proceedings of the 2026 International Conference on Multimedia Retrieval},
  year = {2026},
  pages = {2182--2190},
  doi = {10.1145/3805622.3810787},
}

@inproceedings{zhang2025scene,
  author = {Zhang, Yunzhi and
            Li, Zizhang and
            Zhou, Matt and
            Wu, Shangzhe and
            Wu, Jiajun},
  title = {The {Scene Language}: Representing Scenes with Programs, Words, and Embeddings},
  booktitle = {Proceedings of the IEEE/CVF Conference on Computer Vision and Pattern Recognition},
  year = {2025},
  pages = {24625--24634},
}

@article{jia2025omnisafebench,
  author = {Jia, Xiaojun and
            Liao, Jie and
            Guo, Qi and
            Ma, Teng and
            Qin, Simeng and
            Duan, Ranjie and
            Li, Tianlin and
            Huang, Yihao and
            Zeng, Zhitao and
            Wu, Dongxian and
            others},
  title = {{OmniSafeBench-MM}: A Unified Benchmark and Toolbox for Multimodal Jailbreak
           Attack-Defense Evaluation},
  journal = {arXiv preprint arXiv:2512.06589},
  year = {2025},
}

@inproceedings{jiang-etal-2025-hiddendetect,
  author = {Jiang, Yilei and
            Gao, Xinyan and
            Peng, Tianshuo and
            Tan, Yingshui and
            Zhu, Xiaoyong and
            Zheng, Bo and
            Yue, Xiangyu},
  title = {{HiddenDetect}: Detecting Jailbreak Attacks against Multimodal Large Language Models
           via Monitoring Hidden States},
  booktitle = {Proceedings of the 63rd Annual Meeting of the Association for Computational
               Linguistics (Volume 1: Long Papers)},
  year = {2025},
  pages = {14880--14893},
  doi = {10.18653/v1/2025.acl-long.724},
}

@article{inan2023llama,
  author = {Inan, Hakan and
            Upasani, Kartikeya and
            Chi, Jianfeng and
            Rungta, Rashi and
            Iyer, Krithika and
            Mao, Yuning and
            Tontchev, Michael and
            Hu, Qing and
            Fuller, Brian and
            Testuggine, Davide and
            others},
  title = {{Llama Guard}: {LLM}-based Input-Output Safeguard for Human-{AI} Conversations},
  journal = {arXiv preprint arXiv:2312.06674},
  year = {2023},
}

@misc{nvidia_nemo_guardrails,
  author = {{NVIDIA}},
  title = {Jailbreak Protection},
  howpublished = {\url{https://docs.nvidia.com/nemo/guardrails/configure-guardrails/guardrail-catalog/jailbreak-protection}},
  url = {https://docs.nvidia.com/nemo/guardrails/configure-guardrails/guardrail-catalog/jailbreak-protection},
}

@article{bai2025qwen3,
  author = {Bai, Shuai and
            Cai, Yuxuan and
            Chen, Ruizhe and
            Chen, Keqin and
            Chen, Xionghui and
            Cheng, Zesen and
            Deng, Lianghao and
            Ding, Wei and
            Gao, Chang and
            Ge, Chunjiang and
            others},
  title = {{Qwen3-VL} Technical Report},
  journal = {arXiv preprint arXiv:2511.21631},
  year = {2025},
}

@article{wang2025internvl3,
  author = {Wang, Weiyun and
            Gao, Zhangwei and
            Gu, Lixin and
            Pu, Hengjun and
            Cui, Long and
            Wei, Xingguang and
            Liu, Zhaoyang and
            Jing, Linglin and
            Ye, Shenglong and
            Shao, Jie and
            others},
  title = {{InternVL3.5}: Advancing Open-Source Multimodal Models in Versatility, Reasoning, and
           Efficiency},
  journal = {arXiv preprint arXiv:2508.18265},
  year = {2025},
}

@article{hong2025glm,
  author = {{GLM-V Team} and
            Hong, Wenyi and
            Yu, Wenmeng and
            Gu, Xiaotao and
            Wang, Guo and
            Gan, Guobing and
            Tang, Haomiao and
            Cheng, Jiale and
            Qi, Ji and
            Ji, Junhui and
            others},
  title = {{GLM-4.5V} and {GLM-4.1V-Thinking}: Towards Versatile Multimodal Reasoning with
           Scalable Reinforcement Learning},
  journal = {arXiv preprint arXiv:2507.01006},
  year = {2025},
}

@inproceedings{yu2026minicpm,
  author = {Yu, Tianyu and
            Wang, Zefan and
            Wang, Chongyi and
            Huang, Fuwei and
            Ma, Wenshuo and
            He, Zhihui and
            Cai, Tianchi and
            Chen, Weize and
            Huang, Yuxiang and
            Zhao, Ranchi and
            others},
  title = {{MiniCPM-V 4.5}: Cooking Efficient {MLLMs} via Architecture, Data, and Training
           Recipe},
  booktitle = {Proceedings of the IEEE/CVF Conference on Computer Vision and Pattern Recognition},
  year = {2026},
  pages = {11704--11715},
}

@article{team2025gemma,
  author = {{Gemma Team} and
            Kamath, Aishwarya and
            Ferret, Johan and
            Pathak, Shreya and
            Vieillard, Nino and
            Merhej, Ramona and
            Perrin, Sarah and
            Matejovicova, Tatiana and
            Ram{\'e}, Alexandre and
            Rivi{\`e}re, Morgane and
            others},
  title = {{Gemma 3} Technical Report},
  journal = {arXiv preprint arXiv:2503.19786},
  year = {2025},
}

@misc{openai2026gpt56luna,
  author = {{OpenAI}},
  title = {{GPT-5.6 Luna}},
  howpublished = {\url{https://developers.openai.com/api/docs/models/gpt-5.6-luna}},
  year = {2026},
  url = {https://developers.openai.com/api/docs/models/gpt-5.6-luna},
}

@misc{google2026gemini35flashlite,
  author = {{Google DeepMind}},
  title = {{Gemini 3.5 Flash-Lite}},
  howpublished = {\url{https://deepmind.google/models/model-cards/gemini-3-5-flash-lite/}},
  year = {2026},
  url = {https://deepmind.google/models/model-cards/gemini-3-5-flash-lite/},
}

@misc{anthropic2025claudehaiku45,
  author = {{Anthropic}},
  title = {{Claude Haiku 4.5} System Card},
  howpublished = {\url{https://assets.anthropic.com/m/99128ddd009bdcb/original/Claude-Haiku-4-5-System-Card.pdf}},
  year = {2025},
  url = {https://assets.anthropic.com/m/99128ddd009bdcb/original/Claude-Haiku-4-5-System-Card.pdf},
}

@article{wu2025qwen,
  author = {Wu, Chenfei and
            Li, Jiahao and
            Zhou, Jingren and
            Lin, Junyang and
            Gao, Kaiyuan and
            Yan, Kun and
            Yin, Sheng-ming and
            Bai, Shuai and
            Xu, Xiao and
            Chen, Yilei and
            others},
  title = {{Qwen-Image} Technical Report},
  journal = {arXiv preprint arXiv:2508.02324},
  year = {2025},
}

@inproceedings{liu2024improved,
  author = {Liu, Haotian and
            Li, Chunyuan and
            Li, Yuheng and
            Lee, Yong Jae},
  title = {Improved Baselines with Visual Instruction Tuning},
  booktitle = {Proceedings of the IEEE/CVF Conference on Computer Vision and Pattern Recognition},
  year = {2024},
  pages = {26296--26306},
  doi = {10.1109/CVPR52733.2024.02484},
}

@misc{openai2024gpt4omini,
  author = {{OpenAI}},
  title = {{GPT-4o mini}: Advancing Cost-Efficient Intelligence},
  howpublished = {\url{https://openai.com/index/gpt-4o-mini-advancing-cost-efficient-intelligence/}},
  year = {2024},
  url = {https://openai.com/index/gpt-4o-mini-advancing-cost-efficient-intelligence/},
}

@inproceedings{luo2024jailbreakv,
  author = {Luo, Weidi and
            Ma, Siyuan and
            Liu, Xiaogeng and
            Guo, Xiaoyu and
            Xiao, Chaowei},
  title = {{JailBreakV}: A Benchmark for Assessing the Robustness of MultiModal Large Language
           Models against Jailbreak Attacks},
  booktitle = {First Conference on Language Modeling},
  year = {2024},
}

@inproceedings{zhou2025mssbench,
  author = {Zhou, Kaiwen and
            Liu, Chengzhi and
            Zhao, Xuandong and
            Compalas, Anderson and
            Song, Dawn and
            Wang, Xin Eric},
  title = {Multimodal Situational Safety},
  booktitle = {International Conference on Learning Representations},
  year = {2025},
}

@inproceedings{chen2026mirsafety,
  author = {Chen, Renmiao and
            Lu, Yida and
            Cui, Shiyao and
            Ouyang, Xuan and
            Huang, Victor Shea-Jay and
            Zhang, Shumin and
            Pan, Chengwei and
            Qiu, Han and
            Huang, Minlie},
  title = {The Side Effects of Being Smart: Safety Risks in {MLLM}s' Multi-Image Reasoning},
  booktitle = {Proceedings of the 64th Annual Meeting of the Association for Computational
               Linguistics (Volume 1: Long Papers)},
  year = {2026},
  pages = {36866--36882},
  doi = {10.18653/v1/2026.acl-long.1710},
}

@inproceedings{pmlr-v306-li26jo,
  author = {Li, Pengcheng and
            Zhang, Jie and
            Zhang, Tianwei and
            Qiu, Han and
            Kejun, Zhang and
            Zhang, Weiming and
            Yu, Nenghai and
            Zhou, Wenbo},
  title = {State-Dependent Safety Failures in Multi-Turn Language Model Interaction},
  booktitle = {Proceedings of the International Conference on Machine Learning},
  year = {2026},
  volume = {306},
  pages = {72598--72615},
}

@inproceedings{peng-etal-2026-state2state,
  author = {Peng, Chu-Jun and
            Lin, Zhong-Wei and
            Tseng, Yu-Wen and
            Shuai, Hong-Han and
            Cheng, Wen-Huang},
  title = {{State2State}: Revealing Compositional Safety Failures through State-Transition
           Reasoning},
  booktitle = {Findings of the Association for Computational Linguistics: EMNLP 2026},
  year = {2026},
  pages = {51106--51121},
  doi = {10.18653/2026.findings-emnlp.2379},
}

@inproceedings{xiong2026contextual,
  author = {Xiong, Yuan and
            Miao, Ziqi and
            Li, Lijun and
            Qian, Chen and
            Li, Jie and
            Shao, Jing},
  title = {Contextual Image Attack: How Visual Context Exposes Multimodal Safety Vulnerabilities},
  booktitle = {European Conference on Computer Vision},
  year = {2026},
  pages = {240--257},
  doi = {10.1007/978-3-032-37211-6_14},
}

@inproceedings{pmlr-v306-azulay26a,
  author = {Azulay, Aharon and
            Dubi\'{n}ski, Jan and
            Li, Zhuoyun and
            Mittal, Atharv and
            Gandelsman, Yossi},
  title = {Jailbreaking Vision-Language Models Through the Visual Modality},
  booktitle = {Proceedings of the International Conference on Machine Learning},
  year = {2026},
  volume = {306},
  pages = {4636--4670},
}

@inproceedings{zhao2026vrsa,
  author = {Zhao, Shiji and
            Xiong, Shukun and
            Huang, Yao and
            Yan, Jin and
            Wu, Zhenyu and
            Guan, Jiyang and
            Duan, Ranjie and
            Tao, Jialing and
            Xue, Hui and
            Wei, Xingxing},
  title = {{VRSA}: Jailbreaking Multimodal Large Language Models through Visual Reasoning
           Sequential Attack},
  booktitle = {Proceedings of the IEEE/CVF Conference on Computer Vision and Pattern Recognition
               Findings},
  year = {2026},
  pages = {9412--9421},
}

@inproceedings{schaeffer2025failures,
  author = {Schaeffer, Rylan and
            Valentine, Dan and
            Bailey, Luke and
            Chua, James and
            Eyzaguirre, Cristobal and
            Durante, Zane and
            Benton, Joe and
            Miranda, Brando and
            Sleight, Henry and
            Wang, Tony and
            others},
  title = {Failures to Find Transferable Image Jailbreaks between Vision-Language Models},
  booktitle = {International Conference on Learning Representations},
  year = {2025},
}

@article{xu2026deepseek,
  author = {{DeepSeek-AI} and
            Xu, Anyi and
            Lin, Bangcai and
            Xue, Bing and
            Wang, Bingxuan and
            Xu, Bingzheng and
            Wu, Bochao and
            Zhang, Bowei and
            Lin, Chaofan and
            Dong, Chen and
            others},
  title = {{DeepSeek-V4}: Towards Highly Efficient Million-Token Context Intelligence},
  journal = {arXiv preprint arXiv:2606.19348},
  year = {2026},
}

@article{zeng2025shieldgemma2robusttractable,
  author = {Zeng, Wenjun and
            Kurniawan, Dana and
            Mullins, Ryan and
            Liu, Yuchi and
            Saha, Tamoghna and
            Ike-Njoku, Dirichi and
            Gu, Jindong and
            Song, Yiwen and
            Xu, Cai and
            Zhou, Jingjing and
            others},
  title = {{ShieldGemma 2}: Robust and Tractable Image Content Moderation},
  journal = {arXiv preprint arXiv:2504.01081},
  year = {2025},
}

@article{cohen1960coefficient,
  author = {Cohen, Jacob},
  title = {A Coefficient of Agreement for Nominal Scales},
  journal = {Educational and Psychological Measurement},
  year = {1960},
  volume = {20},
  number = {1},
  pages = {37--46},
  doi = {10.1177/001316446002000104},
}

@article{fleiss1971measuring,
  author = {Fleiss, Joseph L.},
  title = {Measuring Nominal Scale Agreement Among Many Raters},
  journal = {Psychological Bulletin},
  year = {1971},
  volume = {76},
  number = {5},
  pages = {378--382},
  doi = {10.1037/h0031619},
}
